\pdfoutput=1

\documentclass[11pt]{article}

\usepackage[preprint]{acl}

\usepackage{times}
\usepackage{latexsym}

\usepackage[T1]{fontenc}

\usepackage[utf8]{inputenc}

\usepackage{microtype}

\usepackage{inconsolata}

\usepackage{marvosym}
\usepackage{fancyvrb}
\usepackage{xcolor}
\usepackage[most]{tcolorbox}
\usepackage{footmisc}
\usepackage{booktabs}
\usepackage{makecell}
\usepackage{multirow}
\usepackage{siunitx}
\usepackage{listings}
\makeatletter
\newif\ifpreprint
\@ifpackagewith{acl}{preprint}{\preprinttrue}{\preprintfalse}
\makeatother

\title{Can’t Remember Details in Long Documents? You Need Some R\&R}

\author{Devanshu Agrawal\textsuperscript{\Yinyang}, Shang Gao\textsuperscript{\Yinyang}, Martin Gajek \\
  Thomson Reuters \\
  \texttt{\{devanshu.agrawal, shang.gao, martin.gajek\}@thomsonreuters.com} \\
  \Yinyang These authors contributed equally to this work \\}

\begin{document}
\maketitle
\begin{abstract}
Long-context large language models (LLMs) hold promise for tasks such as question-answering (QA) over long documents, 
but they tend to miss important information in the middle of context documents~\citep{liu2023lost}. 
Here, we introduce \emph{R\&R}---a combination of two novel prompt-based methods called \emph{reprompting} and \emph{in-context retrieval} (ICR)---to alleviate this effect in document-based QA. 
In reprompting, we repeat the prompt instructions periodically throughout the context document to remind the LLM of its original task. 
In ICR, rather than instructing the LLM to answer the question directly, we instruct it to retrieve the top $k$ passage numbers most relevant to the given question, which are then used as an abbreviated context in a second QA prompt. 
We test R\&R with GPT-4 Turbo and Claude-2.1 on documents up to 80k tokens in length and observe a 16-point boost in QA accuracy on average. 
Our further analysis suggests that R\&R improves performance on long document-based QA because it reduces the distance between relevant context and the instructions. 
Finally, we show that compared to short-context chunkwise methods, R\&R enables the use of larger chunks that cost fewer LLM calls and output tokens, while minimizing the drop in accuracy.
\end{abstract}

\begin{figure}[t]
\centering
\input{figures/prompt.tex}
\caption{%
Prompt schematic for our method R\&R (see App.~\ref{app:prompts} for full prompts). 
In-context retrieval (blue) abbreviates the document to $k$ passages max (based on the returned IDs), 
and reprompting (red) every $r$ tokens helps to mitigate the ``lost in the middle'' effect. 
Entities in braces are substituted with text, with \{\ldots \} replaced with the instructions nearly verbatim. 
QA is done with the abbreviated context in a second LLM call.}
\label{fig:prompt}
\end{figure}

\section{Introduction}
\label{sec:intro}

Large language models (LLMs) have taken natural language processing (NLP) by storm and are increasingly being incorporated into user-facing applications. 
A user interacts with an LLM via ``prompting'', where the user writes a free-form prompt that instructs the LLM to perform some task, such as answering a question based on a document included in the prompt. 
The prompt, however, is usually limited to a context window with a set maximum number of ``tokens'' (subwords used as the vocabulary of the input language), 
which poses a challenge for tasks such as question-answering (QA) over very long documents. 
Thus, there is great interest in the development of LLMs that support longer and longer context---%
a nontrivial endeavor due to the quadratic complexity of the self-attention mechanism on which LLMs are based.

We have recently witnessed the release of very long-context LLMs such as GPT-4 Turbo and Claude-2.1 supporting context windows of 128k and 200k tokens respectively. 
While these LLMs support long contexts, the quality of their responses (e.g., QA accuracy) tends to deteriorate as the input prompt becomes very long. 
Even at 16k-token context, \citet{liu2023lost} found that an LLM's accuracy on document-based QA significantly drops when relevant context is located in the middle of the document, as opposed to the beginning and end---%
a phenomenon termed the ``lost in the middle'' effect. 
There is thus a demand for techniques that can mitigate this loss of performance, thereby improving the efficacy of long-context LLMs. 
Moreover, as LLMs such as GPT-4 Turbo and Claude-2.1 are proprietary, prompt-based approaches compatible with black-box models are particularly desirable as they can be put into practice immediately.

In this paper, we propose a prompt-based method called \emph{R\&R} to alieviate the lost-in-the-middle effect in long document-based QA (Fig.~\ref{fig:prompt}). 
R\&R, in turn, is a combination of two novel prompt-based methods we call reprompting and in-context retrieval. 
In \emph{reprompting}, the instructions to answer the question are repeated nearly verbatim periodically throughout the context document. 
Our motivation for reprompting is the observation that in the ``lost in the middle'' experiments~\citep{liu2023lost}, the instructions to answer the question appear both before and right after the context document, leading us to wonder: 
Could QA accuracy, in part, be related to the proximity of relevant context in the document to the task instructions? 
Our hypothesis is that repeating the task instructions right before each piece of relevant context in the document will improve QA accuracy. 
Since in practice, however, we do not know the locations of relevant context apriori, we propose to repeat the instructions or ``reprompt'' uniformly.

\emph{In-context retrieval} (ICR) draws inspiration from retrieval-augmented generation (RAG) and recent literature on context abbreviation~\citep{lewis2020retrieval, weston2023system}. 
In ICR, rather than having a long-context LLM answer a question directly, we first prompt the LLM to retrieve some number of passages from the context document that are most relevant to the question. 
We aggregate the retrieved passages into an abbreviated context document, and then do short-context QA in a second LLM call. 
Our underlying hypothesis is that passage retrieval is a simpler task than direct QA, as we are able to prioritize recall over precision. 
R\&R is then a two-call method where we reprompt the instructions for retrieval during the first call in ICR. 
In our experiments, we find R\&R to indeed be beneficial and find evidence to support our motivating hypotheses.

We also evaluate R\&R in the setting of limited-context LLMs that must process long documents in chunks. 
In scenarios where relevant context is concentrated in the document and QA resembles information extraction, a chunkwise approach is likely to provide a strong baseline, thus raising the question: 
Are long-context LLMs with reprompting really better than limited-context LLMs with chunk-based workarounds? 
As a strong baseline, we propose ``chunked ICR'', where we first partition the context document into chunks and then perform ICR within each chunk in independent short-context LLM calls. 
We aggregate the retrieved passages across all chunks into an abbreviated document to do the final QA. 
We expect on extraction-like QA tasks, smaller chunks may facilitate finer retrieval at the cost of more LLM calls, thus introducing an accuracy vs. cost trade-off. 
We find, however, that our method R\&R softens the trade-off by boosting accuracy on larger chunks, often reducing the number of LLM calls and output token usage with little to no drop in accuracy.

Our key contributions are the following: 
(1) We use GPT-4 Turbo and Claude-2.1 to show that reprompting and ICR each independently improve performance on four long-context QA tasks, 
and our combined R\&R approach improves performance further%
\footnote{%
Code for all experiments %
\ifpreprint
is available at: \url{https://github.com/casetext/r-and-r}.
\else
was submitted with this paper as supplementary material.%
\fi}; 
(2) we analyze why reprompting works and show that the proximity between instructions and relevant context plays an important role in performance; 
and (3) we compare long-context QA with chunk-based approaches with shorter context windows and show that R\&R can minimize accuracy loss at higher chunk sizes, thereby reducing LLM usage cost in real-world use cases.

\section{Related Work}
\label{sec:related}

\paragraph{Long context} %
There are a number of avenues in the literature that aim to achieve LLMs that perform well at long context. 
To the extent that long-context LLMs suffer from limited long samples during training, an avenue is to fine-tune LLMs on long-context data so long as one can circumvent complexity challenges~\citep{chen2023longlora, tworkowski2023focused}. 
Alternatively, many avenues propose modifications to the LLM architecture itself, 
including landmark attention~\citep{mohtashami2023landmark}, 
positional interpolation~\citep{chen2023extending} with RoPE embeddings~\citep{su2021roformer}, 
and parallel context windows~\citep{ratner2023parallel}. 
These methods, however, cannot be applied to boost the performance of existing black-box long-context LLMs such as GPT-4 Turbo. 
There is little literature on prompt-based methods intended for such black-box models, 
but one example is ``Found in the middle''~\citep{tang2023found}, 
where permutation-equivariance of LLMs for document reranking is promoted with self-consistency, thereby reducing position bias in the context. 
However, this method likely suffers from complexity issues at very long contexts. 
In contrast to all these methods, we investigate what can be achieved with a set of simple prompt-templates with minimal complexity.

\paragraph{Reprompting} %
There is literature to support the idea of reprompting in addition to the motivation given in Sec.~\ref{sec:intro}. 
For example, attention tends to focus on repeated tokens due in part to how it is trained~\citep{holtzman2020curious, welleck2020neural}, 
and it has been suggested that sycophancy (where the LLM ``seeks approval'' in unwanted ways) is related to this problem of repetition~\citep{roller2021recipes, weston2023system}. 
In reprompting, we seek to exploit this weakness by repeating the task instructions to ensure the LLM does not ``forget'' its goal deep in the middle of the document. 
Indeed, repeating the question has been previously shown to be beneficial~\citep{xu2024re}, 
but in this previous work , the question is only repeated once and almost immediately following its initial statement. 
In contrast, we propose to repeat the question periodically throughout the context document and hypothesize that its mechanism is reducing distance between question and relevant context, as opposed to just repetition.

\paragraph{Retrieval} %
Our inspiration for in-context retrieval (ICR) was the work of \citet{weston2023system}, 
who propose a prompt-based method to reduce sycophancy in QA where an LLM is asked to extract the unbiased parts of the context document. 
In contrast, however, our motivation is to mitigate the ``lost in the middle effect'' and not reduce sycophancy per se. 
Moreover, rather than instructing to do extraction in general, our prompt is more akin to traditional retrieval of chunks. 
Retrieval in general has a long history in NLP (see e.g., the work of \citet{lin2022pretrained} for a review from a text ranking perspective), 
but more recently retrieval and LLMs have come together in retrieval-augmented generation (RAG), where e.g., a long context document can first be chunked, and a retriever is used to effectively abbreviate the document for a downstream LLM~\citep{lewis2020retrieval}. 
With long-context LLMs emerging, \citet{xu2024retrieval} address the important question: 
How do long-context LLMs compare to RAG? 
They find that a 4k-context-window LLM enhanced with RAG is able to achieve performance comparable to that of a 16k-context-window LLM, thus demonstrating the power of chunk-based approaches. 
Here, we ask a similar question but replace RAG with ICR as we are interested in a comparison to what short-context LLMs can achieve natively. 
Moreover, we consider contexts reaching 80k tokens, where the question of long context vs. chunking is unclear.

\section{Our Method}
\label{sec:our}

\subsection{Document-Based QA}
\label{sec:dqa}

In this paper, we focus on the task of document-based QA, where we ask the LLM to answer a question based on the context of an enclosed document. 
For this task, we use a prompt template separated into three top-level blocks: 
First, we enclose the question and the instruction to answer it between \verb|<INSTRUCTIONS>| \ldots \verb|</INSTRUCTIONS>| tags; 
below that, we enclose the document between \verb|<DOCUMENT>| \ldots \verb|</DOCUMENT>| tags; 
finally, we end the prompt by repeating the \verb|INSTRUCTIONS| block almost verbatim to ensure it is the last thing the LLM sees before generating its response. 
This final repetition has been previously done as well~\citep{liu2023lost}, and we regard it as the inspiration for our reprompting method.

Because our full method (R\&R) involves retrieval, we assume the document is separated into ``pages'', 
where each page is enclosed between \verb|<PAGE {p}>| \ldots \verb|</PAGE {p}>| tags with \verb|{p}| replaced with the appropriate page number. 
Here, ``pages'' need not refer to actual pages in the document; 
instead, they could refer to paragraphs, sentences, or any other natural sectional structure of the document. 
In all cases, however, we use the term ``page'' for standardization in this paper.

\subsection{Reprompting}
\label{sec:reprompting}

The lost-in-the-middle effect reveals that LLMs tend to be biased towards the beginning or end of the input prompt or to positions closer to the main instructions enclosed in the prompt. 
To the extent it is the latter, we expect we could mitigate the lost-in-the-middle effect by simply repeating the \verb|INSTRUCTIONS| regularly throughout the document, thereby reducing the positional distance between the instructions and relevant information in the document. 
We thus propose reprompting, where we first construct a block of the form \verb|<INSTRUCTIONS_REMINDER>| Remember, your task is to \ldots \verb|</INSTRUCTIONS_REMINDER>| containing the original instructions almost verbatim. 
For some positive integer $r$, we inject this reminder block approximately every $r$ tokens in the document, outside the \verb|PAGE| blocks. 
QA then proceeds as usual.

\subsection{In-Context Retrieval}
\label{sec:retrieval}

ICR is based on the hypothesis that retrieving information from a document that is relevant to a question is generally a simpler task than answering the question directly, since in the former we prioritize recall over precision. 
We therefore tackle document-based QA with two distinct phases of prompting. 
In phase 1, for some positive integer $k$, we instruct the LLM to retrieve up to $k$ page numbers in the document that are most relevant to the question. 
Then in phase 2, we instruct the LLM to answer the question exactly as described in Sec.~\ref{sec:dqa}, except the document is replaced with the abbreviated version comprising only the pages retrieved in phase 1. 
The number $k$ as well as the pagination level of the document should be jointly chosen such that the abbreviated document comprises a short context that the LLM is able to handle without special prompting techniques. 

Reprompting and ICR can be combined by injecting reminders of the retrieval instructions throughout the document. 
We expect this will help the LLM retrieve relevant pages that may be buried closer to the middle of the document. 
More precisely, in R\&R, we first run phase 1 of ICR with \verb|INSTRUCTIONS_REMINDER| blocks injected every $r$ tokens; 
then, we run phase 2 as usual with no reprompting, as we no longer have long context.

\subsection{Chunking}
\label{sec:chunking}

We propose chunkwise ICR as a strong baseline for extraction-like QA tasks. 
We first split the context document into non-overlapping consecutive chunks of approximately $c$ tokens each, while ensuring that all splits are done outside the \verb|PAGE| blocks. 
In $c$ independent LLM calls, we then perform ICR within each chunk, retrieving up to $ck$ pages in all. 
The retrieved pages constitute the abbreviated document for phase 2 of ICR as described in Sec.~\ref{sec:retrieval}.

If the chunks are large enough, then we can also reprompt with the ICR instructions every $r$ tokens within each chunk---i.e., chunkwise R\&R. 
We hypothesize reprompting within chunks will help to reduce the number of LLM calls required ($c+1$) while minimizing the cost in accuracy.

\section{Experimental Setup}
\label{sec:experimental}

\subsection{Datasets}
\label{sec:datasets}

\paragraph{NQ} %
The NaturalQuestions-Open (NQ) dataset contains historical queries issued to the Google search engine together with human-annotated answers~\citep{kwiatkowski2019natural}. 
We use the same processed version of NQ that was used and made available by \citet{liu2023lost}; 
each example in NQ consists of a question, answer, and a list of passages from Wikipedia ranked by relevance to the question. 
Exactly one passage in the list is annotated as the gold passage containing the answer, while the remaining passages act as ``distracters''. 
For our document-based Q\&A experiments, we take $50$ examples from NQ, 
and we build the document $D$ for each question as a double-linebreak-separated list of the provided passages, such that given positive integers $x \leq d$: 
(1) Each passage is enclosed in \verb|<PAGE {p}>| \ldots \verb|</PAGE {p}>| tags with \verb|{p}| replaced with the appropriate number; 
(2) The gold passage appears approximately $x$ GPT4-tokens into the document; and 
(3) The document is approximately $d$ GPT4-tokens long. 
The distracter passages in the constructed document are still sorted by relevance. 
We call $x$ and $d$ the ``answer position'' and ``document length'' respectively, and we vary their values in our experiments. 
Specifically, we take $d$ to be various multiples of 10k, and we vary $x$ from $0$ to $d$ in increments of 10k. 
With $50$ questions, our NQ dataset has a sample size of
\[ N = 50\left(1 + \frac{d}{10000}\right) = 50 + \frac{d}{200}. \]

\paragraph{SQuAD} %
The Stanford Question-Answering Dataset version 2 (SQuAD) is a dataset of question-answer pairs based on individual paragraphs of Wikipedia articles~\citep{rajpurkar2018know}. 
we take $50$ examples from SQuAD. 
For each question $Q$, we take the accompanying context paragraph $P$ to be the ``gold passage''; 
if $A$ is the Wikipedia article containing $P$, then we take the distracter paragraphs for $Q$ to be paragraphs in SQuAD not contained in $A$. 
Given an answer position $x$ and document length $d$, we then build the context document $D$ for question $Q$ just as for NQ.

\paragraph{HotPotQA} %
This is a multihop Q\&A dataset with questions that require context over multiple Wikipedia articles to answer~\citep{yang2018hotpotqa}. 
It thus allows us to test if our proposed method R\&R is effective when relevant context is scattered throughout a document. 
Each example consists of a question, answer, and a set of paragraphs across Wikipedia articles that together constitute sufficient context to answer the question. 
We take the given paragraphs as the gold passages, and we take paragraphs associated to other questions and unrelated Wikipedia articles to be the distracters. 
Then given a document length $d$, we construct the context document $D$ for a question by taking the distracter passages, inserting the gold passages at regularly spaced intervals, and paginating the passages is in NQ and SQuAD. 
Note the answer position parameter $x$ is inapplicable here, as the relevant contexts are uniformly scatterd throughout the document. 
Since we will not vary $x$ for HotPotQA in our experiments, we take $N = 250$ examples to ensure we have a sufficiently large sample size.

\paragraph{PubMed} %
Our final dataset is a synthetic QA dataset based on biomedical paper abstracts scraped via the PubMed search engine. 
We scraped all abstracts that were published and added to the PubMed index in 2024, and we kept only the ones that were 150-200 GPT4-tokens in length, as this is roughly the range containing the average token length of passages in the NQ, SQuAD, and HotPotQA datasets constructed above. 
Note these abstracts could not have been present in the training data of GPT-4 Turbo and Claude-2.1. 
Given each of the $50$ latest abstracts, we instructed GPT-4 Turbo to write a question that can only be answered based on the given abstract and on no extraneous information, where the answer is either a single-word or short phrase. 
We also stated that the question-answer pair must make sense even if the given abstract were to appear as one paragraph in a much longer document; 
thus, metaquestions such as ``What is the first word in the abstract?'' are excluded. 
For each of the $50$ resulting triplets consisting of question, answer, and context abstract, we took all other abstracts as the distracter passages. 
Given answer position $x$ and document length $d$, we are then able to build the context document $D$ for each question exactly as described for NQ.

In all datasets, we separate every document into \verb|PAGE| blocks that correspond to the natural units out of which the document is built. 
For NQ, SQuAD, and HotPotQA, these correspond to the extracted passages included in the datasets, 
and for PubMed, the natural units are the abstracts themselves.

\subsection{Methods}
\label{sec:methods}

The main methods we test in our experiments are ``Reprompt'' (corresponding to Sec.~\ref{sec:reprompting}) and R\&R. 
Our two key baselines are the following.

\paragraph{Baseline} %
This is a long-context baseline where we just use the standard document-based QA prompt described in Sec.~\ref{sec:dqa}.

\paragraph{Chunking} %
This is a short-context baseline implementing chunkwise ICR as described in Sec.~\ref{sec:chunking}. 
We do not specify the chunk size $c$ used here, as we vary it in our experiments.

In all experiments involving ICR, we retrieve the top $k=5$ most relevant pages, 
as we know $k=5$ is a sufficiently large number for the four datasets under consideration. 
In all experiments involving reprompting, unless stated otherwise, we reprompt every $r=\mbox{10k}$ tokens. 
We provide justification for this choice in Sec.~\ref{sec:analysis}.

\subsection{Evaluation}
\label{sec:evaluation}

We test both GPT-4 Turbo (gpt-4-1106-preview) and Claude-2.1 as our LLM. 
Given a question and document, we compare the LLM-predicted answer $A^{\prime}$ to the ground truth answer $A$ with a symmetric ``fuzzy match'' score, 
which returns $1$ if all unique words in $A$ also appear in $A^{\prime}$ or vice versa (after removing non-alphanumeric non-space characters and capitalization), and $0$ otherwise. 
The fuzzy-match score reported for an entire dataset is the average over all $N$ samples (including both questions and answer positions) in the dataset. 
We find the fuzzy-match score appropriate for our experiments as all answers in our datasets are short sequences of keywords, as opposed to longer open-ended responses; 
it is unlikely for an answer to be correct unless it contains the correct keywords exactly.

\begin{table*}
\centering
\begin{tabular}{ccS[table-format=3.1]SSSSS}
\toprule
\quad & \quad & \multicolumn{3}{c}{{GPT-4 Turbo}} & \multicolumn{3}{c}{{Claude-2.1}} \\
\cmidrule(lr){3-5} \cmidrule(lr){6-8}
\quad & $d$ & {Base} & {Rep} & {R\&R} & {Base} & {Rep} & {R\&R} \\
\midrule
\multirow{4}{*}{NQ} & 10k & 64.0 & & & 50.0 & & \\
 & 20k & 62.0 & 65.3 & 63.3 & 45.3 & 48.7 & 54.7 \\
 & 40k & 57.6 & 63.2 & 60.0 & 43.2 & 52.8 & 43.6 \\
 & 80k & 48.9 & 58.0 & 63.8 & 44.4 & 49.8 & 44.9 \\
\midrule
\multirow{4}{*}{\makecell{SQuAD \\ (SQ)}} & 10k & 96.0 & & & 91.0 & & \\
 & 20k & 94.0 & 94.7 & 94.0 & 73.3 & 84.7 & 92.7 \\
 & 40k & 70.0 & 88.0 & 93.6 & 55.6 & 84.0 & 92.8 \\
 & 80k & 72.0 & 70.4 & 90.9 & 60.4 & 70.4 & 84.4 \\
\midrule
\multirow{4}{*}{\makecell{HotPotQA \\ (HP)}} & 10k & 79.2 & & & 65.6 & & \\
 & 20k & 72.4 & 74.4 & 79.6 & 59.2 & 63.6 & 62.4 \\
 & 40k & 63.6 & 68.0 & 74.0 & 51.2 & 59.6 & 52.8 \\
 & 80k & 50.4 & 54.8 & 61.6 & 41.2 & 56.0 & 53.2 \\
\midrule
\multirow{4}{*}{\makecell{PubMed \\ (PM)}} & 10k & 100.0 & & & 96.0 & & \\
 & 20k & 98.0 & 99.3 & 98.7 & 92.7 & 96.0 & 98.0 \\
 & 40k & 82.0 & 94.8 & 97.2 & 84.4 & 90.4 & 96.4 \\
 & 80k & 66.0 & 77.1 & 95.1 & 75.1 & 88.0 & 95.1 \\
\bottomrule
\end{tabular}
\caption{%
Fuzzy-match scores (\%) of three prompting methods across four document-based QA datasets and four document lengths $d$. 
We omit Reprompt and R\&R at document length 10k as reprompting is not done at this length; 
we list Baseline at 10k, however, as a reference to which longer-context scores can be compared.}
\label{table:scores}
\end{table*}

\section{Results}
\label{sec:results}

\subsection{Main}
\label{sec:main}

In our main set of experiments, we evaluate the benefit of R\&R for document-based QA. 
Table~\ref{table:scores} lists the fuzzy-match scores obtained for each dataset and each long-context method (i.e., excluding chunking) described in Secs.~\ref{sec:datasets}--\ref{sec:methods} and for four different document lengths $d$. 
Reprompt outperforms Baseline almost across the board, 
and R\&R often provides an additional boost especially with GPT-4 Turbo at $d=\mbox{80k}$.%
\footnote{\label{foot:table1}%
We suspect the results for Claude-2.1 on NQ and HotPotQA may be related to the number of pages retrieved, 
as the documents in these datasets contain multiple pages with non-negligible relevance, 
and the results for Claude-2.1 in Table~\ref{table:cvr} are more reasonable as more pages are retrieved chunkwise.} %
The additional cost to run Reprompt is minimal, incurring about $1.15\%$ more input tokens than Baseline at $d=\mbox{80k}$ and no additional output tokens. 
R\&R similarly costs about $1.15\%$ more input tokens compared to Baseline at $d=\mbox{80k}$, but it costs an additional LLM call for the ICR step and incurs about 83 output tokens per sample vs. only 43 output tokens (averaged across all four datasets) with Baseline and Reprompt. 
Nevertheless, our results suggest that our method R\&R can indeed be helpful to extend the context length at which LLMs operate effectively for document-based QA.

\begin{table}
\centering
\begin{tabular}{ccSSSS}
\toprule
\quad & \quad & \multicolumn{2}{c}{{GPT-4 Turbo}} & \multicolumn{2}{c}{{Claude-2.1}} \\
\cmidrule(lr){3-4} \cmidrule(lr){5-6}
\quad & $c$ & {ICR} & {R\&R} & {ICR} & {R\&R} \\
\midrule
\multirow{4}{*}{NQ} & 10k & 54.7 & & 45.6 & \\
 & 20k & 60.7 & 61.3 & 46.7 & 48.7 \\
 & 40k & 60.4 & 60.9 & 43.6 & 47.6 \\
 & 80k & 57.6 & 63.8 & 42.2 & 44.9 \\
\midrule
\multirow{4}{*}{SQ} & 10k & 94.0 & & 95.6 & \\
 & 20k & 94.0 & 94.0 & 79.3 & 91.1 \\
 & 40k & 91.1 & 93.8 & 84.4 & 93.1 \\
 & 80k & 88.7 & 90.9 & 71.8 & 84.4 \\
\midrule
\multirow{4}{*}{HP} & 10k & 78.0 & & 65.6 & \\
 & 20k & 78.8 & 77.6 & 62.0 & 63.2 \\
 & 40k & 70.4 & 76.8 & 58.0 & 57.2 \\
 & 80k & 56.4 & 61.6 & 52.8 & 53.2 \\
\midrule
\multirow{4}{*}{PM} & 10k & 98.0 & & 96.0 & \\
 & 20k & 97.8 & 98.2 & 95.1 & 93.8 \\
 & 40k & 93.1 & 96.7 & 95.8 & 96.2 \\
 & 80k & 89.8 & 95.1 & 94.7 & 95.1 \\
\bottomrule
\end{tabular}
\caption{%
Fuzzy-match scores (\%) of Chunkwise ICR and chunkwise R\&R (see Sec.~\ref{sec:methods}) across four document-based QA datasets at document length $d=\mbox{80k}$ for various chunk sizes $c$. 
We omit R\&R at chunk size 10k as reprompting is not done at this length.}
\label{table:cvr}
\end{table}

To evaluate the benefits of long context and reprompting over short context and a chunk-based approach, we run chunkwise ICR and chunkwise R\&R (the latter is just the former plus reprompting). 
Table~\ref{table:cvr} lists the fuzzy-match scores obtained for each dataset and method. 
We use the max document length $d=\mbox{80k}$ but vary the chunk size $c$ within which ICR and R\&R are run. 
The general trend, more or less, on most of the datasets is that accuracy decreases at larger chunk sizes, as retrieval becomes less accurate with additional filler context.%
\footnote{\label{foot:table2}%
Exceptions to this trend---e.g., ICR on NQ at $c=\mbox{10k}$---are likely due to the effect of chunk size on total number of pages retrieved. 
For more, see Sec.~\ref{sec:limitations}} %
However, reprompting reduces the rate at which accuracy decreases with chunk size and thus may enable larger chunks to be used in practice. 
We understand the significance of this in terms of an accuracy vs. cost trade-off, as smaller chunks cost more LLM calls $m$ (one per chunk, plus QA after aggregation), input tokens, and output tokens (Table~\ref{table:cost}). 
Output tokens, in particular, are costly as their price (\$) is three times that of input tokens for GPT-4 Turbo and LLM run time depends linearly on output tokens. 
Thus, our results suggest that reprompting may help to soften this trade-off by enabling larger chunks that cost fewer LLM calls and output tokens while minimizing loss of accuracy. 
Moreover, reprompting itself costs an insignificant addition of input tokens, 
and in any case this cost is covered by the reduction in input tokens with larger chunks.

\begin{table}
\centering
\begin{tabular}{ccccc}
\toprule
\quad & \quad & \multicolumn{2}{c}{Input tokens} & Output \\
\cmidrule{3-4}
$c$ & $m$ & ICR & R\&R & tokens \\
\midrule
10k & 9 & 82939 & & 322 \\
20k & 5 & 81503 & 81923 & 187 \\
40k & 3 & 80763 & 81334 & 119 \\
80k & 2 & 80369 & 81041 & 84 \\
\bottomrule
\end{tabular}
\caption{%
Number of LLM calls $m$ and average numbers of input tokens and output tokens per question for ICR and R\&R at various chunk sizes $c$ (corresponding to the runs in Table~\ref{table:cvr}). 
ICR and R\&R differ only in the number of input tokens used.}
\label{table:cost}
\end{table}

\subsection{Analysis}
\label{sec:analysis}

\paragraph{Page retrieval} %
Our motivation for ICR is the hypothesis that retrieving the most relevant page(s) from the document is simpler than answering the question directly, as we prioritize recall over precision in the former. 
We test this hypothesis by comparing the task of direct document-based QA with the task of retrieving ``the page most relevant to answering the question''. 
We exclude NQ from this experiment because the ``distracter'' pages appearing closer to the start of each document tend to have non-negligible relevance to the question at hand, and thus there is no clear-cut ``most relevant page''. 
We also exclude HotPotQA because relevant context appears on multiple pages scattered throughout each document. 
On SQuAD and PubMed, however, we see that page retrieval is significantly more accurate than direct question-answering, at the example document length of $d = {40}k$ (Table~\ref{table:page})---thus corroborating our hypothesis.

Observe that the scores for answering in Table~\ref{table:page} are significantly lower than the scores of Baseline at $d = \mbox{40}k$ in Table~\ref{table:scores}. 
The only difference between the two prompts is that in Baseline (as well as in Reprompt and R\&R) we ask the LLM to return the page containing the answer along with the answer itself. 
This simple addition clearly has a significant  positive impact on QA accuracy and illustrates the benefits of page retrieval.

\begin{table}
\centering
\begin{tabular}{cSS}
\toprule
\quad & {Answer} & {Page} \\
\midrule
SQ & 53.6 & 87.6 \\
PM & 68.4 & 94.8 \\
\bottomrule
\end{tabular}
\caption{%
Fuzzy-match scores (\%) for direct document-based QA and exact-match scores (\%) for retrieval of the most relevant page (the page containing the answer) with GPT-4 Turbo at document length $d = {40}k$.}
\label{table:page}
\end{table}

\paragraph{Reprompt rate} %
Here we justify our choice to reprompt every $r=\mbox{10k}$ tokens. 
We test reprompting every 5k, 10k, and 20k tokens at document length $d=\mbox{40k}$, 
and remarkably we find 10k to give the highest QA accuracy across all four datasets (Table~\ref{table:rate}). 
We discuss the likely mechanism underlying this observation next.

\begin{table}
\centering
\begin{tabular}{cSSS}
\toprule
\quad & {5k} & {10k} & {20k} \\
\midrule
NQ & 62.4 & 63.2 & 61.6\\
SQ & 80.8 & 88.0 & 84.4\\
HP & 67.2 & 68.0 & 66.4\\
PM & 94.0 & 94.8 & 92.8\\
\bottomrule
\end{tabular}
\caption{%
Fuzzy-match scores (\%) with reprompting every 5k, 10k, and 20k tokens with GPT-4 Turbo at document length $d=\mbox{40k}$.}
\label{table:rate}
\end{table}

\paragraph{Reprompt position} %
We test the hypothesis described in Sec.~\ref{sec:intro} that reprompting only before relevant context is sufficient to boost accuracy significantly. 
On NQ, SQuAD, and PubMed, we inject just one \verb|INSTRUCTIONS_REMINDER| block before the \verb|PAGE| block marked as containing the ``gold passage''. 
On HotPotQA, we do this for each of the multiple \verb|PAGE| blocks marked as containing relevant context. 
Remarkably, this method results in QA accuracies significantly higher than reprompting uniformly every $r=\mbox{10k}$ tokens on three of the four datasets at the example document length $d=\mbox{40k}$ (Table~\ref{table:reprompts}). 
This finding supports the hypothesis that reprompting works because it reduces the distance between relevant context and at least one copy of the task instructions. 
Furthermore, it elucidates the finding that $r=\mbox{10k}$ is the optimal reprompt rate; 
Recall in our experimental setup, we vary the answer position $x$ in each document in increments of 10k (see Sec.~\ref{sec:datasets} for details); 
in particular, in every sample, every page marked as relevant is an integer multiple of 10k tokens into the document. 
Thus, reprompting every 10k tokens is guaranteed to inject a reminder near the ground truth answer position, while 20k is only guaranteed to do so in half the samples. 
On the other hand, while reprompting every 5k offers the same guarantee as 10k, it is an unnecessarily high reprompt rate and may be introducing contextual noise.

As further corroboration, we test reprompting where we just allude to the original instructions: \verb|<INSTRUCTIONS_REMINDER>| Remember, your task is to follow the instructions under the ``\verb|<INSTRUCTIONS>|'' tag \verb|</INSTRUCTIONS_REMINDER>| 
and find that it performs significantly worse than original reprompting (Table~\ref{table:reprompts}). 
We suspect this test fails because it does not reduce the distance between relevant context and the given question that only appears in the original instructions. 
Finally, we test reprompting where all copies of the reminder block are placed at the beginning of the document 
and find it performs much worse than original reprompting; 
thus, the efficacy of reprompting cannot be attributed to repetition alone.

\begin{table}
\centering
\begin{tabular}{cSSSS}
\toprule
\quad & {Rep} & {\makecell{Tags \\ only}} & {\makecell{At \\ beginning \\ only}} & {\makecell{Before \\ answer \\ only}} \\
\midrule
NQ & 63.2 & 57.6 & 60.0 & 70.4\\
SQ & 88.0 & 81.2 & 70.4 & 81.6\\
HP & 68.0 & 66.4 & 65.2 & 72.0\\
PM & 94.8 & 86.4 & 79.6 & 99.2\\
\bottomrule
\end{tabular}
\caption{%
Fuzzy-match scores (\%) with original and three variations of reprompting at document length $d = \mbox{40k}$. 
``Reprompt'' is taken from Table~\ref{table:scores}; 
in ``Tags only'', the reminder block only refers to the original ``INSTRUCTIONS'' tag; 
in ``At beginning only'', all reminders are placed at the beginning of the document; 
and in ``Before answer only'' only one reminder is placed right before each relevant context page.}
\label{table:reprompts}
\end{table}

\section{Conclusion}
\label{sec:conclusion}

We introduced the prompt-based method R\&R to investigate how far we could push the performance of long-context LLMs on document-based QA. 
We found our nethod to be effective at mitigating the ``lost in the middle effect'' (see App.~\ref{app:triangle} for detailed results), and our analysis suggests that the mechanism underlying reprompting could be the minimization of distance between relevant context and the task instructions. 
For extraction-like QA tasks, chunkwise approaches provide a strong baseline, and indeed R\&R can be performed within chunks. 
Nevertheless, even in this setting, we found reprompting to be beneficial as it often allows larger chunks to be used (thus requiring fewer LLM calls and less token usage) while reducing the drop in accuracy. 
R\&R thus softens the accuracy vs. cost trade-off of chunkwise approaches and enables cost-savings in practical applications where accuracy is paramount.

Future directions of this work are numerous. 
We could combine R\&R with other prompt-based methods to boost performance further; 
e.g., we could ask the LLM to provide a brief justification of each page it retrieves, thus encouraging the LLM to retrieve more wisely. 
Perhaps more interestingly, we could consider ``in-context chunking'', where we ask the LLM to retrieve 5 pages from every 50-page range, for example; 
this could help to temper the accuracy vs. cost trade-off further. 
As a different direction, we could investigate the utility of reprompting on tasks requiring a more wholistic understanding of the document such as summarization, where position bias has also been observed~\citep{ravaut2023position}. 
Finally, while these are only prompt-based methods, understanding their benefits and limitations could help to elucidate the complex behavior of long-context LLMs and could inspire architectural modifications to improve long-context LLMs.

\section{Limitations}
\label{sec:limitations}

While it is clear that R\&R is beneficial compared to the ``Baseline'' QA prompt, the results presented in Tables~\ref{table:scores}--\ref{table:cvr} exhibit some noise. 
We mention potential explanations in footnotes~\ref{foot:table1}--\ref{foot:table2}. 
To elaborate on footnote~\ref{foot:table2}: 
If we ask to retrieve up to $k$ pages from each of $\frac{d}{c}$ chunks at document length $d$, then a max of $\frac{dk}{c}$ pages can be retrieved. 
Thus, in our experimental setup, up to $40$ pages may be retrieved at chunk size $c=\mbox{10k}$. 
In this way, at smaller chunk sizes, the benefits of ICR diminish, leading to a potential decrease in accuracy and in general a more complex accuracy vs. cost trade-off. 
On NQ in particular, pages earlier in the document are semirelevant to the given question, possibly leading to a larger number of pages being retrieved and reducing the benefits of ICR.

We selected GPT-4 Turbo and Claude-2.1 as our LLMs because of their very long contexts, but the catch is their black-box nature. 
This limits our understanding of at least one possible source of noise mentioned above, 
and it limits our investigation into the mechanism of reprompting; 
indeed, with access to a very long-context open-source LLM, we could measure the attention weights to better understand how the LLM responds to reprompting.

Finally, we only consider the task of document-based QA, but we plan to investigate reprompting for other tasks such as summarization in the future.

\ifpreprint
\section*{Acknowledgements}

We would like to thank John Scoville and Pablo Arredondo for their valuable feedback and support.

\else
\fi

\bibliography{references}

\appendix

\section{Prompts}
\label{app:prompts}

Here we list the prompt templates to run our methods. 
In all prompts, \verb|{question}| and \verb|{document}| are replaced with the given question and document text respectively, 
and \verb|{format_instructions}| is replaced with the response format instructions. 
The ``Baseline'' method (as called in Sec.~\ref{sec:results}) runs the following prompt:

\begin{lstlisting}[numbers=none]
<INSTRUCTIONS>
Answer the following question based on the document provided and no additional extraneous information:
{question}

{format_instructions}
</INSTRUCTIONS>	

<DOCUMENT>
{document}
</DOCUMENT>

<INSTRUCTIONS>
Now, answer the following question based on the above document and no additional extraneous information:
{question}

{format_instructions}
</INSTRUCTIONS>
\end{lstlisting}

In the ``Reprompt'' method, before running the above prompt, we first inject the following reminder block every $r$ tokens in the context document:

\begin{lstlisting}[numbers=none]
<INSTRUCTIONS_REMINDER>
Remember, your task is to answer the following question based on this document and no additional extraneous information:
{question}

{format_instructions}</INSTRUCTIONS_REMINDER>
\end{lstlisting}

In ICR, before running the baseline prompt, we first run the following:

\begin{lstlisting}[numbers=none]
<INSTRUCTIONS>
Below is a document that is separated into page numbers. Identify up to 5 page numbers in the document that are most relevant to the following question: 
{question}
	
{format_instructions}
</INSTRUCTIONS>

<DOCUMENT>
{document}
</DOCUMENT>

<INSTRUCTIONS>
Now, identify up to 5 page numbers in the document that are most relevant to the following question. 
{question}

{format_instructions}
</INSTRUCTIONS>'''
\end{lstlisting}

Finally, in ``R\&R'', we inject the following reminder block every $r$ tokens into the context document before running the above ICR prompt:

\begin{lstlisting}[numbers=none]
<INSTRUCTIONS_REMINDER>
Remember, your task is to identify up to 5 page numbers in the document that are most relevant to the following question: 
{question}

{format_instructions}
</INSTRUCTIONS_REMINDER>
\end{lstlisting}

\section{Effect of Answer Position}
\label{app:triangle}

Here we present the results in Table~\ref{table:scores} in greater detail, broken down by answer position 
(Recall from Sec.~\ref{sec:datasets} that for each question in a dataset, we vary the position $x$ of the answer in the context document in increments of 10k). 
We omit HotPotQA as we do not vary the positions of the multiple pieces of relevant context in this dataset. 
Tables~\ref{table:triangle:nq}-\ref{table:triangle:pubmed} list the results for NQ, SQuAD, and PubMed respectively. 
On all three datasets, we observe the ``lost in the middle'' effect in Baseline, particularly at document length $d=\mbox{80k}$. 
On the other hand, we observe consistently that ICR+Rep mitigates this effect significantly.

\begin{table*}
\centering
\begin{tabular}{ccSSSSSSSSS}
\toprule
\quad & $d$ & {0k} & {10k} & {20k} & {30k} & {40k} & {50k} & {60k} & {70k} & {80k} \\
\midrule
\multirow{4}{*}{Base} & 10k & 60.0 & 68.0 & & & & & & & \\
 & 20k & 60.0 & 56.0 & 70.0 & & & & & & \\
 & 40k & 60.0 & 50.0 & 50.0 & 56.0 & 72.0 & & & & \\
 & 80k & 56.0 & 36.0 & 38.0 & 42.0 & 46.0 & 48.0 & 48.0 & 52.0 & 74.0 \\
\midrule
\multirow{4}{*}{Rep} & 10k & & & & & & & & & \\
 & 20k & 60.0 & 64.0 & 72.0 & & & & & & \\
 & 40k & 60.0 & 56.0 & 64.0 & 64.0 & 72.0 & & & & \\
 & 80k & 54.0 & 38.0 & 52.0 & 58.0 & 60.0 & 62.0 & 60.0 & 64.0 & 74.0 \\
\midrule
\multirow{4}{*}{R\&R} & 10k & & & & & & & & & \\
 & 20k & 64.0 & 64.0 & 62.0 & & & & & & \\
 & 40k & 62.0 & 58.0 & 58.0 & 60.0 & 62.0 & & & & \\
 & 80k & 62.0 & 62.0 & 60.0 & 58.0 & 66.0 & 68.0 & 64.0 & 68.0 & 66.0 \\
\bottomrule
\end{tabular}
\caption{%
Fuzzy-match scores on NQ from Table~\ref{table:scores} broken down by answer position. 
We vary the answre position in increments of 10k tokens within the document length $d$.}
\label{table:triangle:nq}
\end{table*}

\begin{table*}
\centering
\begin{tabular}{ccSSSSSSSSS}
\toprule
\quad & $d$ & {0k} & {10k} & {20k} & {30k} & {40k} & {50k} & {60k} & {70k} & {80k} \\
\midrule
\multirow{4}{*}{Base} & 10k & 96.0 & 96.0 & & & & & & & \\
 & 20k & 96.0 & 90.0 & 96.0 & & & & & & \\
 & 40k & 46.0 & 34.0 & 88.0 & 88.0 & 94.0 & & & & \\
 & 80k & 68.0 & 26.0 & 60.0 & 74.0 & 70.0 & 88.0 & 84.0 & 88.0 & 90.0 \\
\midrule
\multirow{4}{*}{Rep} & 10k & & & & & & & & & \\
 & 20k & 96.0 & 94.0 & 94.0 & & & & & & \\
 & 40k & 84.0 & 76.0 & 94.0 & 92.0 & 94.0 & & & & \\
 & 80k & 54.0 & 24.0 & 46.0 & 78.0 & 78.0 & 90.0 & 82.0 & 88.0 & 94.0 \\
\midrule
\multirow{4}{*}{R\&R} & 10k & & & & & & & & & \\
 & 20k & 94.0 & 94.0 & 94.0 & & & & & & \\
 & 40k & 94.0 & 94.0 & 94.0 & 92.0 & 94.0 & & & & \\
 & 80k & 92.0 & 78.0 & 86.0 & 92.0 & 94.0 & 94.0 & 94.0 & 94.0 & 94.0 \\
\bottomrule
\end{tabular}
\caption{%
Fuzzy-match scores on SQuAD from Table~\ref{table:scores} broken down by answer position. 
We vary the answre position in increments of 10k tokens within the document length $d$.}
\label{table:triangle:squad}
\end{table*}

\begin{table*}
\centering
\begin{tabular}{ccSSSSSSSSS}
\toprule
\quad & $d$ & {0k} & {10k} & {20k} & {30k} & {40k} & {50k} & {60k} & {70k} & {80k} \\
\midrule
\multirow{4}{*}{Base} & 10k & 100.0 & 100.0 & & & & & & & \\
 & 20k & 100.0 & 96.0 & 98.0 & & & & & & \\
 & 40k & 100.0 & 84.0 & 66.0 & 64.0 & 96.0 & & & & \\
 & 80k & 100.0 & 76.0 & 46.0 & 52.0 & 52.0 & 56.0 & 60.0 & 58.0 & 94.0 \\
\midrule
\multirow{4}{*}{Rep} & 10k & & & & & & & & & \\
 & 20k & 100.0 & 100.0 & 98.0 & & & & & & \\
 & 40k & 100.0 & 94.0 & 96.0 & 88.0 & 96.0 & & & & \\
 & 80k & 96.0 & 70.0 & 84.0 & 66.0 & 66.0 & 76.0 & 70.0 & 74.0 & 92.0 \\
\midrule
\multirow{4}{*}{R\&R} & 10k & & & & & & & & & \\
 & 20k & 100.0 & 98.0 & 98.0 & & & & & & \\
 & 40k & 100.0 & 96.0 & 98.0 & 96.0 & 96.0 & & & & \\
 & 80k & 100.0 & 94.0 & 96.0 & 94.0 & 94.0 & 94.0 & 94.0 & 96.0 & 94.0 \\
\bottomrule
\end{tabular}
\caption{%
Fuzzy-match scores on PubMed from Table~\ref{table:scores} broken down by answer position. 
We vary the answre position in increments of 10k tokens within the document length $d$.}
\label{table:triangle:pubmed}
\end{table*}

\end{document}